\documentclass[sn-mathphys-num]{sn-jnl}

\usepackage{graphicx}%
\usepackage{multirow}%
\usepackage{amsmath,amssymb,amsfonts}%
\usepackage{amsthm}%
\usepackage{mathrsfs}%
\usepackage[title]{appendix}%
\usepackage{xcolor}%
\usepackage{textcomp}%
\usepackage{manyfoot}%
\usepackage{booktabs}%
\usepackage{algorithm}%
\usepackage{algorithmicx}%
\usepackage{algpseudocode}%
\usepackage{listings}%

\usepackage{tabularx}
\usepackage{tabu}

\theoremstyle{thmstyleone}%
\theoremstyle{thmstyletwo}%

\theoremstyle{thmstylethree}%

\begin{document}

\title[Article Title]{MPT: A Large-scale Multi-Phytoplankton Tracking Benchmark}

\author[1]{\fnm{Yang} \sur{Yu}} 

\author*[1]{\fnm{Yuezun} \sur{Li}}\email{liyuezun@ouc.edu.cn}

\author[2]{\fnm{Xin} \sur{Sun}}\email{sunxin1984@ieee.org}

\author*[1]{\fnm{Junyu} \sur{Dong}}\email{dongjunyu@ouc.edu.cn}

\affil[1]{\orgdiv{College of Computer Science and Technology}, \orgname{Ocean University of China}, \orgaddress{\city{Qingdao}, \country{China}}}

\affil[2]{\orgdiv{Faculty of Data Science}, \orgname{City University of Macau}, \orgaddress{\city{Macau}, \country{China}}}



\abstract{Phytoplankton are a crucial component of aquatic ecosystems, and effective monitoring of them can provide valuable insights into ocean environments and ecosystem changes. Traditional phytoplankton monitoring methods are often complex and lack timely analysis. Therefore, deep learning algorithms offer a promising approach for automated phytoplankton monitoring. However, the lack of large-scale, high-quality training samples has become a major bottleneck in advancing phytoplankton tracking.
In this paper, we propose a challenging benchmark dataset, Multiple Phytoplankton Tracking (MPT), which covers diverse background information and variations in motion during observation. The dataset includes 27 species of phytoplankton and zooplankton, 14 different backgrounds to simulate diverse and complex underwater environments, and a total of 140 videos. To enable accurate real-time observation of phytoplankton, we introduce a multi-object tracking method, Deviation-Corrected Multi-Scale Feature Fusion Tracker(DSFT), which addresses issues such as focus shifts during tracking and the loss of small target information when computing frame-to-frame similarity.
Specifically, we introduce an additional feature extractor to predict the residuals of the standard feature extractor's output, and compute multi-scale frame-to-frame similarity based on features from different layers of the extractor. Extensive experiments on the MPT have demonstrated the validity of the dataset and the superiority of DSFT in tracking phytoplankton, providing an effective solution for phytoplankton monitoring.}



\maketitle

\begin{figure}[!t]
  \centering
  \scalebox{1.0}{
  \includegraphics[width=1\linewidth]{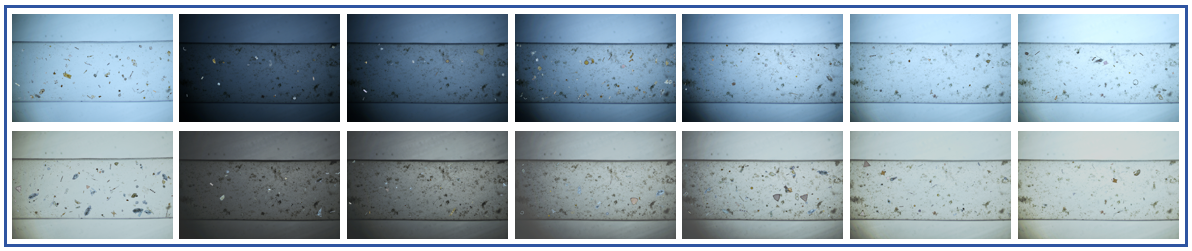}}
    \vspace{-0.1cm}
  \caption{\small A dataset with 14 different background images. There are variations in brightness and impurity levels under white and blue backgrounds.}
  \label{fig:dataset}
\end{figure}

\section{Introduction}\label{sec1}

Phytoplankton is an ecological concept referring to tiny plants that live in water through a planktonic lifestyle, typically microalgae~\cite{matsunaga2005marine}. They are the primary producers in aquatic ecosystems and the main source of dissolved oxygen in water bodies~\cite{chen2018microalgae}. Monitoring phytoplankton is an effective way to understand changes in the marine environment and ecosystems. Therefore, it is of great importance to monitor phytoplankton for maintaining the Earth's ecological balance, protecting water resources, and determining the productivity of aquatic environments.~\cite{ahmad2021role}

Traditional phytoplankton monitoring methods often rely on sampling techniques, where water samples are filtered through membranes with specific pore sizes to capture the phytoplankton.~\cite{alvarez2011effectively} The membranes are then treated for transparency and observed manually under a microscope. This method requires significant human and material resources and cannot provide real-time observation. Collecting large-scale data to train deep neural networks and using these networks for automatic identification can solve these problems. However, existing multi-object tracking methods often struggle to accurately track phytoplankton in complex underwater environments. This is due to two main reasons: the lack of high-quality, large-scale phytoplankton video datasets, which limits the ability to fully train algorithms, and the fact that current multi-object tracking algorithms are not well-suited to the specific characteristics of the underwater environment and phytoplankton.

Creating a comprehensive phytoplankton dataset presents significant challenges due to equipment and environmental limitations. While it is relatively straightforward to collect water samples, obtaining a wide range of phytoplankton species and capturing them in large-scale video data is much more difficult. Phytoplankton are highly diverse, and collecting enough species to represent their variety in natural environments requires extensive fieldwork, specialized equipment, and ideal environmental conditions, which are not always available.

Moreover, even when samples are obtained, the process of capturing high-quality video footage that accurately reflects the dynamic behavior of phytoplankton can be complex. This includes factors such as maintaining stable conditions in the laboratory, ensuring proper microscopy settings, and recording long sequences at high frame rates to observe movement. Due to these difficulties, most existing plankton datasets primarily consist of static image data, which lacks the temporal information necessary for tracking and analyzing plankton motion.

\begin{figure}[!t]
  \centering
  \scalebox{1.0}{
  \includegraphics[width=0.7\linewidth]{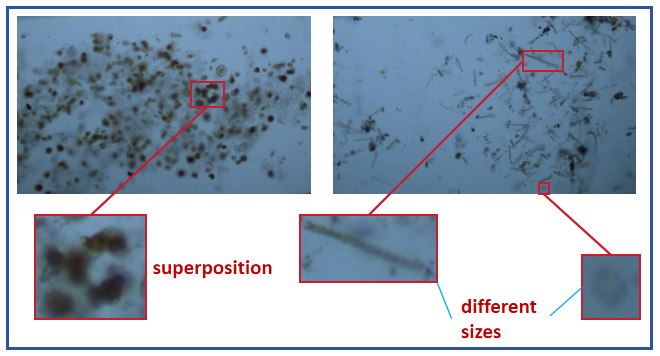}}
    \vspace{-0.1cm}
  \caption{\small In real-world scenarios, overlapping phytoplankton individuals can lead to difficulties in identification (left image); different species of phytoplankton also exhibit significant size differences (right image).}
  \label{fig:example}
\end{figure}

Even video datasets like PMOT2023~\cite{yu2023pmot2023}, though a valuable resource, are limited in both sample density and scale. The video data available often includes only a small number of species or frames, and the lack of variety and richness in the samples hinders comprehensive analysis and generalization in multi-object tracking tasks. As a result, there's a growing need for larger, more diverse video datasets that can better represent the complexity of plankton behavior and their environmental interactions.

Without access to large-scale, high-quality video datasets, it becomes challenging to fully develop and test algorithms that can perform effectively across diverse aquatic conditions. To overcome this limitation, we have constructed MPT, a synthetic large-scale phytoplankton video benchmark dataset (refer Fig.\ref{fig:dataset}), specifically designed to facilitate the training and evaluation of multi-object tracking algorithms under various environmental backgrounds.

The MPT dataset was created using samples primarily sourced from the coastal waters of the Yellow Sea, where a wide range of plankton species can be found. To enhance the dataset’s diversity and better simulate real-world conditions, we incorporated different video backgrounds and environmental variables. Specifically, we designed 14 distinct background images, each featuring variations in lighting conditions, background colors (white and blue tones), and different levels of impurities to mimic natural aquatic environments more closely. This approach ensures that tracking algorithms trained on MPT can generalize across a wide range of environmental scenarios.

Moreover, MPT includes video sequences featuring 27 different plankton species, providing a rich variety of data points. In total, the dataset consists of 140 high-resolution videos, capturing both the diversity of plankton and the complexities of their interactions in different environments. By simulating such a wide range of conditions, the MPT dataset ensures that the tracking algorithms developed using this resource can handle the variability and unpredictability of real-world aquatic environments. This synthetic yet realistic dataset fills a crucial gap in the field, offering researchers a valuable tool for advancing the state of the art in phytoplankton tracking and monitoring.

Based on the MPT dataset, we have developed a multi-object tracking framework named DSFT, designed for real-time tracking of phytoplankton. Currently, there are two main shortcomings when using multi-object tracking algorithms to process phytoplankton data(refer Fig.\ref{fig:example}):
\begin{enumerate}
    \item Phytoplankton often resemble the natural aquatic environment in appearance. When multiple phytoplankton overlap with each other or overlap with impurities, it can cause the algorithm's attention to shift inappropriately. 

    \item When tracking is performed using a similarity matrix between consecutive frames, the algorithm typically achieves high accuracy. However, this approach has a drawback: using only the deepest feature maps from the previous and current frames to calculate similarity can result in the loss of small object information, which is especially problematic in phytoplankton tracking, as there are significant size differences among different species. 
\end{enumerate}

Due to these limitations, existing multi-object tracking algorithms rarely perform well in phytoplankton tracking. To address these two issues, we propose the following solutions.
\textbf{1)} We introduce the Deviation Correction Method (DCM), adding an auxiliary backbone that takes the original image and normal feature maps as input. The output is fused with the original feature maps to correct deviations, ensuring that the algorithm focuses its attention on individual targets.
\textbf{2)} We propose the Multi-scale Feature Similarity Fusion (MFSF). During feature extraction, we extract shallow, intermediate, and deep features to correspond to small, medium, and large objects, respectively. We then calculate the similarity between consecutive frames at each feature level to prevent the loss of information from small and medium objects.

In this paper, we first introduce the related work in the field of phytoplankton, tracking datasets, and tracking algorithms. And we provide a detailed explanation of the creation process and methodology of the MPT dataset. The design of the DSFT algorithm is explained chapter by chapter, covering the overall workflow and the computational principles of each module. The effectiveness of the algorithm is demonstrated in the experimental section.

The contributions of this paper are threefold:
\begin{enumerate}
    \item We have established a large-scale video benchmark dataset MPT (27 species and 140 videos) in the field of phytoplankton.

    \item We propose a phytoplankton-specific tracking framework, DSFT, which effectively addresses the limitations of traditional tracking methods in this domain.

    \item We validated the robustness of the MPT dataset and the effectiveness of the DSFT framework through various experiments.
    
\end{enumerate}


\begin{figure}[!t]
  \centering
  \scalebox{1.0}{
  \includegraphics[width=1\linewidth]{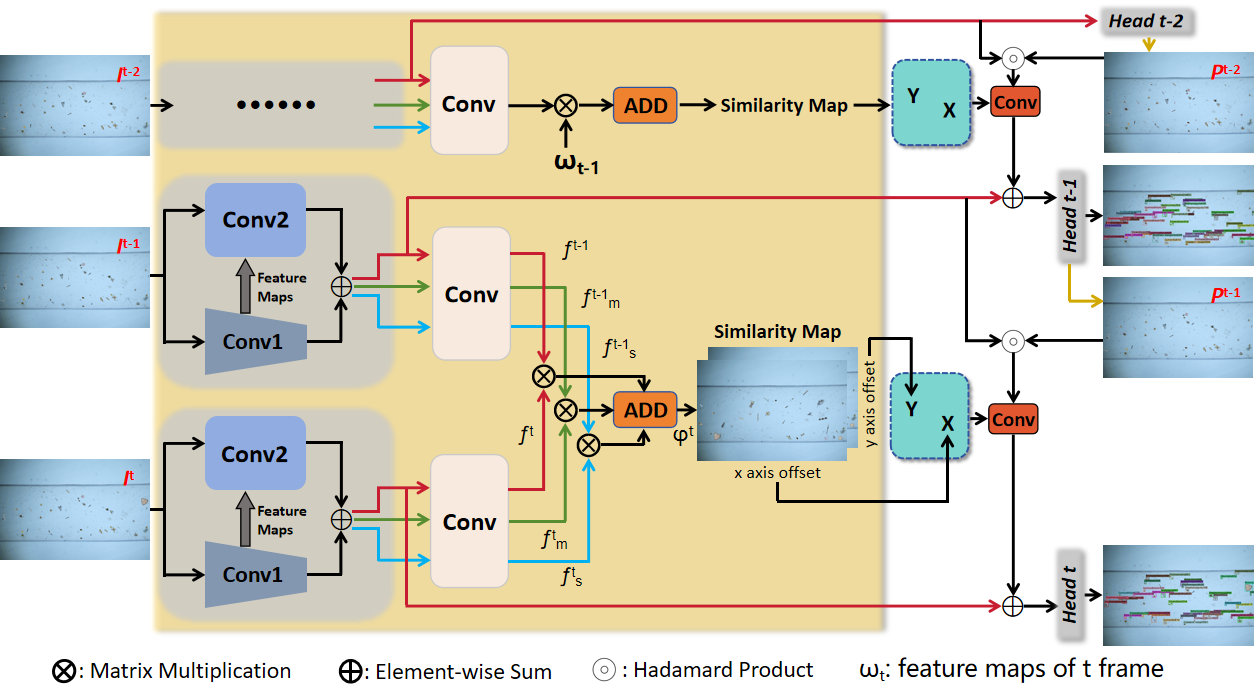}}
    \vspace{-0.1cm}
  \caption{\small Overview of DSFT. The lines of different colors in the figure are not overlapping at the connection points. The red arrow represents the deepest feature map output by Conv1, the green arrow represents the intermediate feature map output by Conv1, and the blue arrow represents the shallow feature map output by Conv1. The gold arrow indicates the detection result generated from the head of the previous frame, which contains only positional information and no class information.}
  \label{fig:overview}
\end{figure}

\section{Related Work}\label{sec2}

\subsection{Phytoplankton}\label{subsec2}
Marine phytoplankton, these microscopic plant-like organisms, play an indispensable role in ocean ecosystems~\cite{anderson1995hydrogen}. They are a group of tiny algae and prokaryotes that float freely in the surface and mid-water layers of the ocean, comprising eight major groups: cyanobacteria, green algae, diatoms, dinoflagellates, golden algae, yellow-green algae, cryptophytes, and euglenoids. Phytoplankton are crucial to the Earth's carbon cycle and climate regulation~\cite{sellner1987phytoplankton}. Through photosynthesis, they absorb carbon dioxide and release oxygen, not only providing a food source for marine ecosystems but also supplying essential oxygen to marine life. Therefore, these microscopic organisms form the foundation of the marine food chain and have a profound impact on global climate change and ecological balance.

The population dynamics, biomass, and responses of marine phytoplankton to environmental changes are key areas of marine science research~\cite{fenchel1988marine}. By continuously monitoring the distribution and density of these microorganisms, researchers can assess the health of marine ecosystems, predict the potential impacts of climate change on marine ecology, and effectively manage fisheries resources. When phytoplankton experience excessive growth, such as during harmful algal blooms (e.g., red tides), they can disrupt marine ecosystems, affect water quality, and degrade the quality of seafood, ultimately posing negative consequences for human activities.

\subsection{Phytoplankton Monitoring}\label{subsec2}
Marine phytoplankton play an irreplaceable role in the global carbon cycle~\cite{tett2008use}, maintaining the marine food chain, and supporting the development of aquaculture, making their monitoring critically important~\cite{boyce2015patterns,reynolds1984phytoplankton,irwin2015phytoplankton}. Although traditional monitoring methods—such as microscopic observation, spectroscopic and fluorescence analysis, remote sensing, and biochemical analysis—have provided valuable data and insights, these techniques often fall short of accurately and in real-time monitoring the population density and distribution of marine phytoplankton~\cite{smayda1997harmful}. In this field, the emergence of FlowCAM technology has brought revolutionary advances to phytoplankton monitoring. FlowCAM, a cutting-edge device that integrates flow cytometry and microscopic imaging technology, is designed for the automatic detection and identification of particles in water samples. This technology can provide detailed information on the quantity, size, shape, and, in some cases, classification of various particles in a sample~\cite{2006Use}.

However, although the above methods provide strong support for laboratory based quantitative analysis of water samples, they still require significant human and material resources. This presents a challenge: the lack of an in-situ monitoring method and corresponding dataset for real-time observation of phytoplankton population density and distribution in natural water bodies.

\subsection{Dataset}\label{subsec2}

In the field of Multi-Object Tracking (MOT), commonly used public datasets include MOT16, MOT17, MOT20~\cite{milan2016mot16}, and DanceTrack~\cite{sun2022dancetrack}. The commonly used planktonic datasets include WHOI Plankton~\cite{orenstein2015whoi}, PMID2019~\cite{li2020developing}, and PMOT2023~\cite{yu2023pmot2023}.

MOT16 and MOT17 are widely used benchmark datasets in the MOT domain, designed to provide a variety of complex scenarios and organized by the MOT Challenge. MOT17, compared to MOT16, introduces more sequences and richer annotations, including detection results generated by different detectors, allowing researchers to test their algorithms under more diverse conditions. MOT20, released after MOT16 and MOT17 by the MOT Challenge, focuses on extremely crowded scenes. Compared to previous datasets, MOT20 provides much higher crowd density. DanceTrack is a relatively new dataset that presents a unique challenge to the MOT field: tracking multiple dancers in dance videos. Unlike traditional surveillance videos, the videos in DanceTrack feature complex human movements and significant occlusions between individuals. This not only demands the algorithm's ability to dynamically predict movements but also requires it to accurately understand human poses and motion patterns.

The WHOI Plankton dataset, created by the Woods Hole Oceanographic Institution, is a large-scale collection of plankton images designed for classification tasks. This dataset comprises over 3.4 million high-quality images of phytoplankton, annotated by experts, and covers 70 different species categories. However, while this dataset is highly valuable for identifying and classifying various types of plankton, it only offers image-level annotations. As a result, it is not suitable for tasks like object tracking, where continuous observation of movement is required.

In 2019, the PMID2019 dataset was introduced as a phytoplankton detection benchmark. Although useful in detection tasks, the dataset consists primarily of stained images, which differ significantly from real-time, in-situ plankton observations. Furthermore, similar to the WHOI Plankton dataset, PMID2019 also provides image-level data and lacks the temporal information necessary for tracking tasks, limiting its application in dynamic environments where motion is critical.

PMOT2023, unveiled in 2023, marked a major step forward by introducing the first dataset specifically designed for multi-object tracking of plankton. Unlike previous datasets, PMOT2023 focuses on simulating plankton movement in controlled environments, such as flow tubes observed under a microscope. By offering video sequences of plankton in motion, it effectively fills a gap in the available resources for plankton observation. This synthetic dataset provides crucial video data for tracking multiple plankton species simultaneously, making it a valuable resource for studying their behavior and movement patterns over time.

\subsection{Algorithm}\label{subsec2}
Current tracking methods are primarily developed for ground-based environments, concentrating on general objects like vehicles and pedestrians~\cite{sun2016visual,zhu2021rgbt,yao2018semantics,li2018learning}. In the context of multi-object tracking within video sequences, three critical components are essential for determining object trajectories: object extraction, temporal association, and motion prediction. Based on the tracking workflow, these methods can be categorized into two main types: offline tracking and online tracking.

Offline tracking makes use of data from future frames and typically represents the problem using a graph model to find a globally optimal solution~\cite{jordan2004graphical,1708000,7041176}. However, this approach is not practical for real-world applications since it depends on information from frames that have not yet occurred. In contrast, online multi-object tracking focuses on establishing correspondences between current detections and existing object trajectories based solely on the data at hand~\cite{9127935,8853333,8540936}. This method mandates that the tracking decisions for each frame are based only on the information from the current and preceding frames, without using current data to modify past results. Consequently, this paper centers on the online tracking framework.


\section{Dataset Establishment}\label{sec3}

\subsection{Sample Analysis}\label{subsec2}

In this study, we employed a high-resolution microscope equipped with a 4K resolution camera to conduct detailed observations and sampling of phytoplankton from coastal areas in the Yellow Sea. Phytoplankton samples were meticulously collected using a plankton net to ensure their representativeness and accuracy. These samples were subsequently examined under the microscope, and high-resolution images were captured at a resolution of 2720x1824 pixels. This approach facilitated comprehensive analysis of the morphological characteristics and detailed features of the phytoplankton specimens.

The sampling period spanned the entire year, with a particular focus on months of heightened phytoplankton density, such as September and October. As the diversity of phytoplankton species in coastal regions of the Yellow Sea is somewhat limited, additional phytoplankton images were sourced from publicly available online repositories to supplement our dataset. Careful selection criteria were applied during image collection to ensure backgrounds were consistent with those observed in our samples, thereby minimizing potential issues arising from background color discrepancies that could affect dataset quality.

In constructing the MPT dataset, we meticulously documented 27 distinct types of phytoplankton species observed during our study. These include Ceratium furca, Gymnodinium, Ceratium, Anabaena, Copepoda, Copepod nauplii, Coscinodiscus, Chaetoceros, Odontella, Leptocylindru, Paralia sulcata, Melosira, Pseudo-nitzschia, Asterionella, Guinardia, Protoperidinium, Pleurosigma, Bellerochea, Thalassiosira, Stephanopyxis, Ditylum, Entomoneis, Akashiwo sanguinea, Rhizosolenia, Biddulphia, Triceratium, Hemiaulus. Each species was meticulously documented to ensure comprehensive coverage and accuracy in our dataset for subsequent tracking and analysis studies.

\subsection{Production Method}\label{subsec2}



The background images in our dataset are composed of two distinct color schemes: blue and white, with seven images for each color, resulting in a total of 14 unique background images. For each of these backgrounds, we applied varying levels of impurity density and brightness to simulate different aquatic environments. This variation allows for the creation of more realistic scenarios and increases the robustness of the tracking algorithms being developed and tested.

For each background image, we generated 10 sequences of consecutive image frames, with each sequence saved as a video at 25 frames per second. To simulate the natural movement of phytoplankton in real-world water environments, we employed jittering and rotation mechanisms on each phytoplankton sample incorporated into the video. These movements helped mimic the natural flow, rotation, and drift that phytoplankton typically exhibit, thus enhancing the realism of the dataset. The jittering and rotational variance add dynamic complexity to the sequences, contributing to a more challenging and comprehensive dataset.

To ensure a balanced and rational distribution of data across the dataset, the types and quantities of phytoplankton included in each video sequence were randomly selected. Furthermore, we varied the total number of frames, as well as the degree of jitter and the movement speed of the phytoplankton, all within a predetermined range. This approach provided a rich diversity in the sequences, allowing the dataset to capture a wide range of motion patterns and environmental conditions, thereby supporting the development of more generalized and adaptable tracking algorithms for real-world applications.

\subsection{Advantages of MPT}\label{subsec2}

The MPT dataset provides significant advantages over existing phytoplankton tracking datasets, making it an invaluable resource for advancing multi-object tracking in underwater environments. One of its key strengths is the large scale and diversity of data. The dataset includes 140 high-resolution video sequences featuring 27 different phytoplankton species, offering extensive coverage for robust algorithm training.
MPT focuses on video sequences rather than static images, enabling continuous tracking of phytoplankton over time. These high-quality videos, recorded in 4K resolution, allow for detailed analysis of small-scale objects and provide a realistic representation of phytoplankton movement—critical for MOT algorithms. The dataset simulates real-world aquatic environments with 14 different backgrounds, incorporating variations in impurity density, lighting, and brightness. This diversity ensures that tracking algorithms can handle dynamic, real-world scenarios.

In addition, MPT includes phytoplankton species of various sizes, posing challenges for traditional tracking algorithms. However, by offering high-resolution images and simulating realistic motions through jittering and speed variation, the dataset ensures that algorithms trained on it can adapt to different object scales and movements. The dataset is designed for real-time tracking, with sequences saved at 25 frames per second, making it suitable for applications in marine ecological monitoring and underwater research.
MPT’s synthetic and scalable nature makes it a flexible tool for research. The dataset not only supports tracking tasks but also addresses detection tasks, bridging the gap left by previous datasets focused solely on classification.


\section{Methods}\label{sec4}

Traditional multi-object tracking algorithms face two main challenges when tracking plankton: first, when there is overlap or partial overlap between individuals, the algorithm's focus can shift inappropriately; second, when using the similarity matrix between consecutive frames for tracking analysis, the algorithm may lose information on smaller objects. To address these issues, we describe an online tracker for plankton called DSFT (refer Fig.\ref{fig:overview}).
First, we propose the DCM, which corrects feature map biases and ensures the algorithm focuses on the individual being tracked. In addition, we introduce the MFSF, which emphasizes the connection between plankton of different sizes across frames, effectively enhancing the algorithm’s ability to detect small plankton.

\subsection{Overall Process}\label{subsec2}
Our method's overall process is inspired by TraDes~\cite{trades}. Given an input variable $\mathcal{I}_t$, it is first passed through a feature extraction module, where the DCM is applied to extract bias-corrected feature maps, denoted as $f^t$. During this process, intermediate features $f^t_m$ and shallow features $f^t_s$ are also extracted from the backbone and corrected. The three levels of features from both the previous and current frames are then multiplied and fused to establish the correlation between them. This correlation allows us to accurately predict the motion offsets of phytoplankton of different sizes. Finally, the motion offsets are passed through the convolutional and head networks for the final prediction.

\begin{figure}[!t]
  \centering
  \scalebox{1.0}{
  \includegraphics[width=0.5\linewidth]{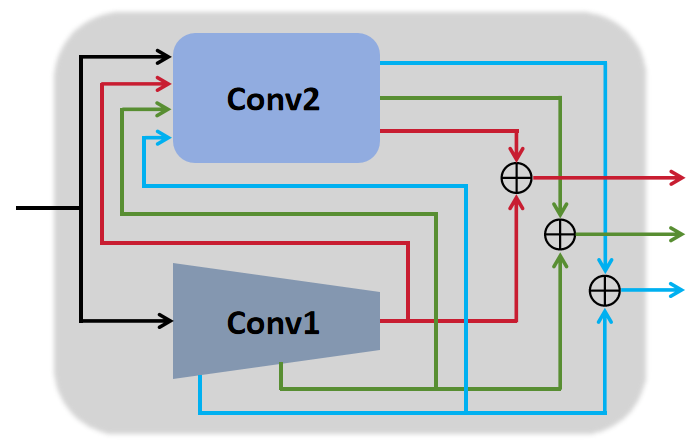}}
    \vspace{-0.1cm}
  \caption{\small Flowchart of the DCM section. In the figure, the connection points of lines with different colors do not overlap. The red, green, and blue arrows represent the outputs from different layers of the Conv1 module.}
  \label{fig:model1}
\end{figure}

\subsection{Deviation Correction Method}\label{subsec2}
Plankton often exhibit appearances similar to their aquatic environment. When plankton overlap or partially overlap with other individuals or with surrounding debris, the algorithm's focus may shift inappropriately. This characteristic makes traditional multi-object trackers unsuitable for tracking plankton. Inspired by DINO-Tracker, our method introduces an additional feature extractor to predict the residual error of the primary feature extractor's results. We use this residual to correct the bias in the main feature map. (refer Fig.\ref{fig:model1})

Specifically, during the feature extraction stage, the current frame is passed through the Conv1 module to generate the main feature map $f^t$. The main feature map and the current frame are then passed through Conv2 to obtain a residual, which represents the bias in the extracted features. We add this residual to the main feature map, resulting in a corrected feature map that mitigates the bias. Then we pass this corrected feature map into the convolution module to calculate Similarity Map.

\subsection{Multi-scale Feature Similarity Fusion}\label{subsec2}
Different species of plankton often exhibit significant size differences, and when the algorithm focuses on larger individuals, it tends to overlook the features of smaller ones. In multi-object tracking algorithms, using a similarity matrix between consecutive frames typically yields good tracking results. However, calculating this matrix using only the deepest feature maps can result in the loss of information about smaller objects.

To address this, we propose the MFSF. Inspired by GeneralTrack, this method aims to emphasize features of smaller individuals while still focusing on larger ones. To achieve this, we extract additional shallow features $f^t_s$ and mid-level features $f^t_m$ during the feature extraction process, corresponding to smaller and medium-sized objects. The shallow and mid-level features from consecutive frames are multiplied to generate their respective similarity matrices, and the results are fused.

\begin{equation}
  \varphi^t = (f^{t-1} \times f^t) + (f^{t-1}_m \times f^t_m) + (f^{t-1}_s \times f^t_s)
\end{equation}

The fused feature map highlights the similarity information of plankton of different sizes across frames.


\begin{table}[!t]
\caption{The performance of various methods on the MPT dataset. When the metric values are less than or equal to 0, we replace them with ‘-’.}\label{compare}%
\begin{tabular}{@{}c llllllll@{}}
\toprule
Method & MOTA$\uparrow$ & IDF1$\uparrow$ & IDs$\downarrow$ & FP$\downarrow$ & FN$\downarrow$ \\
\midrule
Sort (ICIP'16)\cite{sort} & 6.2 & 10.0 & 924 & \textbf{128} & 15307 \\
DeepSort (ICIP'17)\cite{deepsort} & - & 7.6 & 800 & 153 & 17345 \\
TraDeS (CVPR'21)\cite{trades} & 30.2 & 30.4 & 2037 & 1355 & 8786 \\
BotSort (Arxiv'22)\cite{aharon2022botsort} & 11.6 & 25.7 & \textbf{198} & 988 & 14242 \\
ByteTrack (ECCV'22)\cite{zhang2022bytetrack} & 15.6 & 36.7 & 202 & 2172 & 12360 \\
UAVMot (CVPR'22)\cite{uavmot} & 12.3 & 30.3 & 392 & 1783 & 13126 \\
StrongSort (TMM'23)\cite{du2023strongsort} & 9.5 & 12.3 & 283 & 1642 & 13924 \\
UCMCTrack (AAAI'24)\cite{yi2024ucmc} & 13.4 & 8.6 & 622 & 1406 & 12937 \\
BoostTrack (MVA'24)\cite{stanojevic2024boostTrack} & 22.3 & 32.2 & 305 & 2274 & 9764 \\
TLTDMOT (CVPR'24)\cite{chen2024delving} & 18.7 & 24.1 & 340 & 1518 & 14626 \\
\midrule
\textbf{\textit{DSFT}} & \textbf{53.6} & \textbf{46.2} & 2352 & 2001 & \textbf{3737} \\
\botrule
\end{tabular}
\label{table_compare}
\end{table}

\section{Experiments}\label{sec5}

In this section, we introduce the evaluation metrics, experimental parameter design, and comparison methods, and validate the overall experimental performance as well as the effectiveness of the proposed modules.

\subsection{Experimental Settings}\label{subsec2}

\subsubsection{Evaluation Metrics}\label{subsubsec2}

We utilize the CLEAR-MOT metrics~\cite{bernardin2008evaluating} to assess tracking performance. This metric includes several indicators, such as MOTA, IDF1, IDs, FP, and FN. Among them, MOTA and IDF1 are the most important indicator. The explanation of each metric is as follows:  
\begin{enumerate} 
    \item[-] MOTA is a comprehensive metric that evaluates the overall effectiveness of a tracker across the dataset; 
    \item[-] IDF1 assesses the tracker's capacity to maintain consistent identity labels throughout the entire sequence; 
    \item[-] False Positives (FP) are instances where the tracking algorithm falsely identifies a target that does not exist; 
    \item[-] False Negatives (FN) occur when the tracking algorithm misses a target that is actually present; 
    \item[-] Identity switches (IDs) refer to cases where the tracker incorrectly alters the identity of a target during tracking. 
\end{enumerate}

\subsubsection{Implementation Details}\label{subsubsec2}
During the training process, we applied data augmentation to each batch of data with a certain probability. The input size of the images was uniformly adjusted to 860 * 640, and the learning rate was reduced at the 40th and 50th epochs. We utilized the Conv1 network for primary feature extraction and employed the Conv2 network to extract the residuals from the main feature maps. The model was trained for 60 epochs on a single 4080 GPU card with a batch size of 2, and an initial learning rate of 2.0e-4. During the inference stage, the tracking confidence threshold was set to 0.4, and the input image size was set to 640 * 960. Inputs were processed in pairs of two consecutive frames simultaneously. In the comparative methods of the tracking-by-detection category, we uniformly employed CenterNet as the detector, with DLA34 serving as the backbone for CenterNet. The remaining parts were set to the default configuration of TraDeS. Conv1 represents DLA34, and Conv2 represents Delta-DINO.

\begin{table}[t]
\setlength{\tabcolsep}{12pt}
\renewcommand{\arraystretch}{1.0}
\normalsize
\centering
\caption{\small b1-b7 and w1-w7 represent videos under different backgrounds, where 'b' stands for 'blue' and 'w' stands for 'white,' corresponding to the labels in Figure 1. Different numbers represent different impurity concentrations and brightness conditions. 'Average' refers to the overall score on our test set.}
\begin{tabular}{l|c c}
\hline
\textbf{DSFT} & \textbf{MOTA (\%)} & \textbf{IDF1 (\%)} \\
\hline
b1 & 50.8 & 44.6 \\
b2 & 54.0 & 43.1 \\
b3 & 51.4 & 37.7 \\
b4 & 63.2 & 60.6 \\
b5 & 58.6 & 53.4 \\
b6 & 58.0 & 51.2 \\
b7 & 67.3 & 62.5 \\
w1 & 59.6 & 49.2 \\
w2 & 59.6 & 50.0 \\
w3 & 54.8 & 46.0 \\
w4 & 51.6 & 36.1 \\
w5 & 44.5 & 30.3 \\
w6 & 58.1 & 53.9 \\
w7 & 52.3 & 47.2 \\
\hline
\textbf{Average} & 53.6 & 46.2 \\
\hline
\end{tabular}
\label{DSFT}
\end{table}

\subsubsection{Compared Methods}\label{subsubsec2}
We have chosen several different methods for comparison, and the specific methods are as follows:
\begin{enumerate}
    \item[-] SORT (ICIP'16)~\cite{sort}: SORT is a straightforward and real-time multi-object tracking method that applies a Kalman filter combined with the Hungarian algorithm for data association. This significantly improves both tracking speed and accuracy.
    \item[-] DeepSORT (ICIP'17)~\cite{deepsort}: DeepSORT builds on SORT by incorporating ReID(Re-Identification)~\cite{luo2019bag} and appearance features from detection boxes. It employs a Matching Cascade strategy to minimize the occurrence of target ID switches.
    \item[-] TraDeS (CVPR'21)~\cite{trades}: TraDeS is an end-to-end joint detection and tracking model that leverages tracking signals to support detection. It infers tracking offsets from cost metrics and uses these to propagate prior target features, improving detection and segmentation of current targets.
    \item[-] BotSORT (Arxiv'22)~\cite{aharon2022botsort}: BotSORT combines the strengths of both motion and appearance cues, while compensating for camera movement and refining the Kalman filter’s state vector for more precise tracking results.
    \item[-] ByteTrack (ECCV'22)~\cite{zhang2022bytetrack}: ByteTrack applies a two-step matching strategy. First, it matches high-confidence detection boxes with tracks, then associates lower-confidence boxes with unmatched tracks from the first step. It handles occlusions without relying on a ReID model, using only a Kalman filter and the Hungarian algorithm.
    \item[-] UAVMOT (CVPR'22)~\cite{uavmot}: UAVMOT is designed for tracking from drone perspectives, building on FairMOT. It includes an ID feature update module, an adaptive motion filter, and a gradient balanced focal loss, each addressing drone footage challenges like complex motion and improving frame-to-frame ReID feature consistency.
    \item[-] StrongSORT (TMM'23)~\cite{du2023strongsort}: StrongSORT enhances DeepSORT by refining the feature extractor, introducing an inertia term for smoother feature updates, utilizing a Kalman filter tailored for non-linear motion, and incorporating motion data into the cost matrix.
    \item[-] UCMCTrack (AAAI'24)~\cite{yi2024ucmc}: UCMCTrack addresses challenges caused by erratic camera motion by linking the motion of objects with their positions relative to the ground. It uses a mapped Mahalanobis distance instead of IoU to measure the similarity between objects and their previous trajectories.
    \item[-] BoostTrack (MVA'24)~\cite{stanojevic2024boostTrack}: BoostTrack addresses unreliable detections and ID switches by introducing a confidence score for detection tracklets. It scales the similarity metric using this score, and combines Mahalanobis distance with shape similarity to improve tracking accuracy and reduce IoU-induced ambiguities.
    \item[-] TLTDMOT (CVPR'24)~\cite{chen2024delving}: TLTDMOT targets the long-tail distribution problem in multi-object tracking by implementing two data augmentation techniques: Static Camera Viewpoint Augmentation and Dynamic Camera Viewpoint Augmentation. It also features a Group Softmax module for re-identification to handle the distribution imbalance.
\end{enumerate}

These techniques encompass both tracking-by-detection and end-to-end tracking strategies. Their areas of improvement differ, with some concentrating on refining the filtering of detection boxes, while others focus on mitigating the effects of camera shake. 
All of these methods are trained and evaluated under the same configuration as our approach.

We compared the aforementioned methods with DSFT on the MPT dataset, and the experimental results are shown in Table~\ref{compare}.

We tested the DSFT algorithm on the MPT dataset, which includes varying background colors, different impurity densities, and varying brightness conditions. The specific scores for each part are shown in Table~\ref{DSFT}.

\begin{table}[t]
\setlength{\tabcolsep}{9pt}
\normalsize
\centering
\renewcommand{\arraystretch}{0.9}
\caption{\small The effectiveness of the two modules under different embedding scenarios, validating the efficacy of each module.}
\begin{tabular}{l|cc|cc}
\hline
Scheme & DCM & MFSF & MOTA$\uparrow$ & IDF1$\uparrow$ \\
\hline
1 Baseline &  &  & 30.2 & 30.4 \\
2 & $\checkmark$ &  & 46.4 & 37.1 \\
3 &  & $\checkmark$ & 41.5 & 39.8 \\
4 \textbf{\textit{DSFT}} & $\checkmark$ & $\checkmark$ & \textbf{53.6} & \textbf{46.2} \\
\hline
\end{tabular}
\label{ablation}
\end{table}

\subsection{Ablation Study}\label{subsec2}

In this section, we validate the effectiveness of the proposed DSFT through an ablation study. All experiments were conducted on the MPT dataset. To ensure a fair evaluation of each component's performance, the training and testing details remain consistent with those described in the implementation section. When testing specific modules, no modifications were made to the remaining components.

To assess the effectiveness of the two proposed methods, we conducted three sets of experiments: adding only DCM, adding only MFSF, and adding both them under identical conditions. As shown in Table~\ref{ablation}, the performance significantly improved with the addition of each module compared to the baseline alone. When both methods were integrated, the MOTA and IDF1 scores increased by 23.4\% and 15.8\%, respectively, compared to the baseline.


\section{Conclusion}\label{sec6}

In this paper, we present a large-scale plankton tracking dataset, MPT, and develop a multi-object tracking method specifically designed for plankton, named DSFT. MPT contains 140 high-definition videos and 27 species of plankton, incorporating 14 distinct background images that vary in brightness, impurity density, and color. DSFT is a real-time multi-object tracking framework that offers an automated and timely solution for monitoring plankton. To address two key limitations of traditional tracking methods—namely, improper focus shifts when objects overlap and the potential loss of small object information during tracking analysis—we propose two solutions: the DCM to correct feature bias and the MFSF to enhance the representation of small objects. These methods improve both the performance and reliability of the tracking algorithm.

Extensive experiments on MPT demonstrate the validity of the dataset and the superiority of DSFT in plankton tracking. This work not only fills the gap in existing video datasets for plankton but also provides a tailored solution for plankton tracking, laying a foundation for future advancements in marine ecological monitoring and scientific exploration.


\section*{Declarations}


\subsection*{Funding}\label{subsec2}

This work was supported in part by Sanya Science and Technology Special Fund 2022KJCX92.

\subsection*{Conflicts of interest/Competing interests}\label{subsec2}

Not applicable.

\subsection*{Data Availability Statement}\label{subsec2}

Our code and dataset can be found at the following website:

\noindent https://github.com/chyangyu/MPT.git

\subsection*{Authors' contributions}\label{subsec2}

Yang Yu is responsible for the experimental part, dataset production, and paper writing. Yuezun Li assisted in the work and assisted in revising the paper. Xin Sun and Junyu Dong assisted in revising the paper.

\subsection*{Ethics approval}\label{subsec2}

Not applicable.

\subsection*{Consent to participate}\label{subsec2}

Not applicable.

\subsection*{Consent for publication}\label{subsec2}

Not applicable.

\subsection*{Acknowledgments}\label{subsec2}

Not applicable






\bibliography{reference}

\end{document}